\documentclass[letterpaper, 10 pt, conference]{ieeeconf}  % Comment this line out if you need a4paper
\IEEEoverridecommandlockouts                              % This command is only needed if 
\usepackage[utf8]{inputenc} % allow utf-8 input
\usepackage{amsmath,amsfonts}       % blackboard math symbols
\usepackage{amsthm}
\usepackage[T1]{fontenc}    % use 8-bit T1 fonts
\usepackage{hyperref}       % hyperlinks
\usepackage{url}            % simple URL typesetting
\usepackage{booktabs}       % professional-quality tables
\usepackage{nicefrac}       % compact symbols for 1/2, etc.
\usepackage{microtype}      % microtypography
\usepackage{xcolor}         % colors
\usepackage{algorithm,algpseudocode}
\usepackage[textsize=tiny]{todonotes}
\newtheorem{theorem}{Theorem}[section]
\newtheorem{proposition}[theorem]{Proposition}

\theoremstyle{definition}

\newtheorem{example}[theorem]{Example}
\usepackage{cite}

\title{Extremum-Seeking Action Selection for Accelerating Policy Optimization}

\author{Ya-Chien Chang and Sicun Gao}

\begin{document}
\maketitle

\begin{abstract}
Reinforcement learning for control over continuous spaces typically uses high-entropy stochastic policies, such as Gaussian distributions, for local exploration and estimating policy gradient to optimize performance. Many robotic control problems deal with complex unstable dynamics, where applying actions that are off the feasible control manifolds can quickly lead to undesirable divergence. In such cases, most samples taken from the ambient action space generate low-value trajectories that hardly contribute to policy improvement, resulting in slow or failed learning. We propose to improve action selection in this model-free RL setting by introducing additional adaptive control steps based on Extremum-Seeking Control (ESC). On each action sampled from stochastic policies, we apply sinusoidal perturbations and query for estimated Q-values as the response signal. Based on ESC, we then dynamically improve the sampled actions to be closer to nearby optima before applying them to the environment. Our methods can be easily added in standard policy optimization to improve learning efficiency, which we demonstrate in various control learning environments. 
\end{abstract}

\section{Introduction}

Deep reinforcement learning offers a promising solution for challenging control problems in robotics~\cite{PengCZLTL20,kalashnikov18a,wang2019autonomous,li2021reinforcement,KaufmannLR0K020,chang2021stabilizing,so2023solving,ganai2023iterative}. It turns controller synthesis into stochastic optimization problems in the parameter space of expressive control policies. In such policy optimization process, each vector of parameters defines a stochastic policy, from which we sample trajectories to estimate the policy gradient direction over the parameters, for local improvement of the policy towards higher overall performance. A wide range of techniques have been developed to ensure reliable gradient estimation and policy improvement~\cite{trpo,ddpg,ppo,td3,sac}. However, it is still often observed that in many control problems, policy optimization can fail to make progress towards desirable performance~\cite{engstrom2020implementation,hsu2020revisiting,ibarz2021train}. 

A common design in policy optimization over continuous spaces that has not been challenged much is that the policies map states to Gaussian distributions over the action spaces. 
Sampling using such high-entropy stochastic policies enables exploration of actions, which is generally important in RL. 
However, many robotic control problems deal with complex unstable dynamics, where applying actions that are off the feasible control manifolds can quickly lead to undesirable divergence. Consequently, in such cases, most samples taken from the ambient action space generate low-value trajectories that hardly contribute to policy improvement, resulting in slow or failed learning. For instance, consider the problem of controlling the thrusts of a quadrotor for path tracking. Successful control requires symmetry in the thrusts of the four propellers. Exploration of actions under a Gaussian distribution over the four actions will most likely violate such constraints, and the quadrotor can quickly lose balance and fall off, making it hard to progress in learning. Such problems have led to much pre-processing and manual tuning in the practical use of RL~\cite{engstrom2020implementation,hsu2020revisiting,hu2020learning,ibarz2021train,gupta2022unpacking}, such as normalization of states and observations, reward engineering, and customized design of state and action spaces such that the feasible control becomes easier to sample with Gaussian distributions. 

\begin{figure}[t!] 
    \centering
\includegraphics[width=\linewidth]{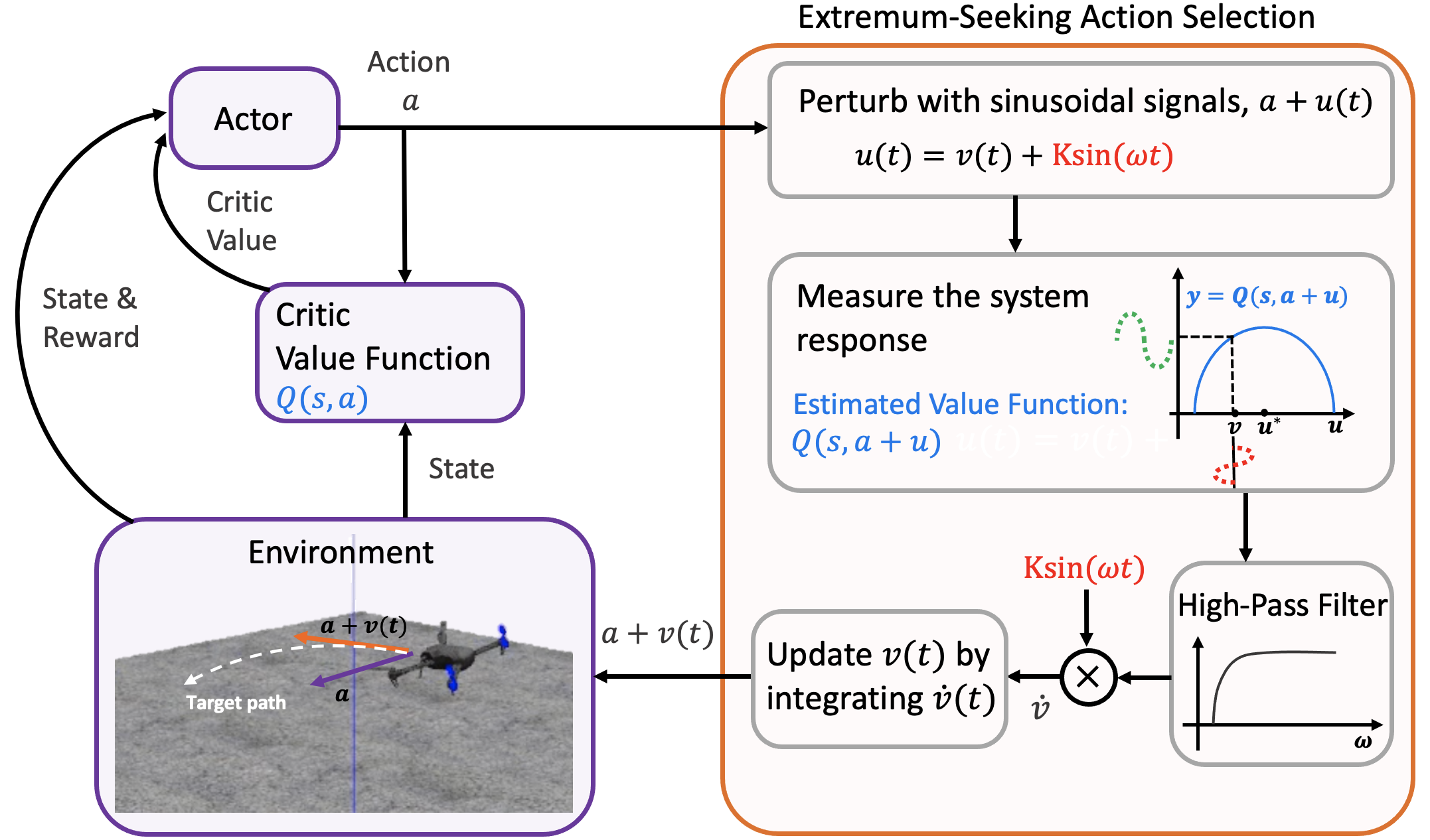}
    \caption{Diagram for Extremum-Seeking Action Selection (ESA) in the RL setting. We use Extremum-Seeking Control (ESC) strategies to improve the quality of exploratory actions, which reduces the sampling of low-value trajectories and accelerates policy optimization.}
    \label{fig:overview} 
\end{figure}

In this paper, we propose a general model-free approach for improving the quality of exploratory action samples for accelerating policy optimization. 
We utilize the methods of Extremum-Seeking Control (ESC), an adaptive feedback control strategy that performs real-time optimization of system performance~\cite{escbook,zhang2011extremum,nevsic2009extremum}. ESC injects periodic perturbation signals in the control input to a system, and formulates feedback control laws to maintain dynamic estimations that adapt to the system response. It can improve the performance of the system by tracking its local optimum without analytic knowledge of the underlying dynamics. 
We do not directly change the practice of using Gaussian distributions as stochastic policies, but adapt ESC to the RL context to improve the quality of sampled trajectories. After each action is sampled from stochastic policies in the standard way, we apply sinusoidal perturbations and query for estimated Q-values as the response signal (Figure~\ref{fig:overview}). Based on ESC, we then dynamically improve the sampled actions to be closer to nearby optima before applying them to the environment. 

We will demonstrate how the ESC components can be easily added in standard policy optimization algorithms such as PPO~\cite{ppo} and SAC~\cite{sac} to improve learning efficiency, without requesting any additional models or oracles that are often needed in DPI approaches~\cite{DBLP:conf/nips/0002GBB18}.

In Section~\ref{sec:simple}, we first use simple model-free optimization settings to demonstrate why ESC can achieve better sample quality and efficiency than policy gradient methods. Then in Section~\ref{sec:esa}, we describe our Extreme-Seeking Action Selection (ESA) method for policy optimization that uses ESC to improve action selection and reduces the sampling of low-value trajectories, to accelerate the learning process. In Section~\ref{sec:exp}, we evaluate the proposed methods in various continuous control environments in MuJoCo~\cite{MuJoCo}, as well as a high-fidelity simulation environment for quadrotor control. We show the benefits of adding ESA in PPO and SAC, as well as other approaches such as adding parameter noise. We perform various ablation studies to show basic strategies for hyperparameter tuning in ESA, which is important in balancing between fast adaptive improvement on each sample and the conservative updates by policy gradient. 

\vspace{2mm}
\noindent\textbf{Related Work.} 
The integration of adaptive control and reinforcement learning has been investigated in recent work~\cite{khan2012reinforcement,richardsadaptive,westenbroek2020adaptive, annaswamy2023integration,wang2023reinforcement}. Typically the focus is on 
using both methods for better control design in systems with specific structures such as control-affineness. Under such assumptions, strong theoretical guarantees can be obtained on stability and asymptotic convergence, which significantly improves standard RL approaches. 
The work in~\cite{richardsadaptive,westenbroek2020adaptive, annaswamy2023integration} uses RL approaches to first train a nearly optimal controller that is subsequently integrated into adaptive control methods to ensure global boundedness with asymptotic stability. 
We aim to propose a new direction for using methods from adaptive control to RL that focuses on improving learning efficiency in the general model-free setting. The dynamics perspectives and frequency-domain techniques from ESC can avoid the well-known challenge of maintaining complex distributions for sampling high-value trajectories, and allow us to improve each sampled action in a manner similar to real-time model-free control. 

It is important to note that what we propose here is not just an exploration strategy, but a way for enhancing the quality of exploratory samples to accelerate the learning process in challenging control problems. Exploration is a major topic in RL~\cite{Survey_exploration}. Over continuous spaces, a widely used technique is to introduce noise into the policy action space~\cite{wawrzynski2015control,ddpg,td3, eberhard2022pink, ruckstiess2010exploring} or parameter space~\cite{conti2018improving, fortunato2018noisy,DBLP:conf/iclr/PlappertHDSC0AA18,mahajan2019maven}. We introduce sinusoidal perturbations based on ESC design to enable the use of frequency-domain techniques for improving each action sample, which is different from further injecting random noise. We will further discuss the difference and compare the performance with the exploration through noise injection approaches in the experiments. 

\section{Preliminaries}
\label{sec:simple}

\subsection{Extreme-Seeking Control}\label{sec:esc}

Extremum-seeking control (ESC) is a model-free adaptive control method for adjusting inputs to a system to dynamically track its locally optimal performance. Here we explain it in the simplest setting of the system being an unknown but static objective function, in which case ESC can be viewed as a zeroth-order optimization method. 

Consider a continuous objective function $J(u): \mathbb{R}^n\rightarrow \mathbb{R}$. Starting from an initial input $u(0)\in\mathbb{R}^n$, ESC designs a control law for tracking a nearby local optimum $u^*$ of $J$ without accessing the analytic form of $J$. The core idea is that by injecting periodic perturbations on the input $u(t)$ and observing the change in $J(u(t))$, we can perform frequency-domain analysis to exploit the derivative information of $J$. To achieve this, the ESC introduces an additional 
estimation variables $v(t)\in\mathbb{R}^n$ that aim to converge to $u^*$. The feedback control law on $u(t)$ and $v(t)$ are designed as:
\begin{eqnarray}
u(t) &=& K\sin(\omega t) + v(t)\label{first}\\
\dot v(t) &=& -\alpha\mathbb{L}[\sin(\omega t)\mathbb{H}[J(u(t))]]\label{eq:v}
\end{eqnarray}
where $\mathbb{H}$ represents a high-pass filter, and $\mathbb{L}$ represents a low-pass filter. Note that $v(t)$ is updated through its time derivative $\dot v(t)$ in Equation (\ref{eq:v}), from initial condition $v(0)=u(0)$. Importantly, the actual input to the system $u(t)$ in Equation (\ref{first}) is always oscillatory, and it is the estimation vector $v(t)$ that will converge to $u^*$. $K\in \mathbb{R}^+$ is the magnitude of the sinusoidal perturbation in $u(t)$, and $\alpha\in\mathbb{R}$ is a signed learning rate parameter. 

At a high-level, the design uses the control input $u(t)$ to probe the system response $J(u(t))$. Then for the estimation $v(t)$, we apply high-pass filtering to the response $\mathbb{H}[J(u(t))]$, and then demodulate with $\sin(\omega t)\mathbb{H}[J(u(t))]$, and finally use a low-pass filter $\mathbb{L}[\sin(\omega t)\mathbb{H}[J(u(t))]]$. After these steps, we will be able to make use of second-order properties of the objective $J$ for dynamically updating $v(t)$ to approach $u^*$. For example, in the one-dimensional case, we can show (more details of the derivation are provided in the Appendix section):
\begin{multline}\label{eq:third}
\frac{\mathrm{d}}{\mathrm{d}t} (v(t)-u^*)=-\dot v(t)= \alpha K\mathbb L[\sin(\omega t)\mathbb{H}[J(u(t))]]\\ 
= -\frac{1}{2}\alpha KJ''(u^*)\cdot (v(t)-u^*)
\end{multline}
Thus, if $J$ is convex around $u^*$ (the concave case can be handled by changing the sign of $\alpha$), then $-\frac{1}{2}\alpha KJ''(u^*)$ is real and negative, and the error 
between $v(t)$ and $u^*$ follows exponentially-stable linear dynamics that quickly converges to zero. Namely, the estimation $v(t)$ will converge to the optimum $u^*$ of the objective $J$. Importantly, although the analysis uses $J''(u^*)$, it never needs to be known or estimated, since the algorithm only needs to iteratively update $u(t)$ and $v(t)$ according to Equation (\ref{first}) and (\ref{eq:v}). In general, ESC methods ensure the following:
\begin{proposition}[Convergence of ESC~\cite{escbook}]
With appropriate sinusoidal perturbations and the corresponding filters, the estimation $v(t)$ exponentially converges to a local optimum $u^*$ of the objective function $J$ in a neighborhood of $v(0)$.
\end{proposition}
Although the method sketched above considers a static objective, the power of ESC methods lies in its ability of dealing with stochastic objectives. The idea is that as long as the plant dynamics has lower-frequency than the perturbation, the high-pass filter will remove the intrinsic frequency of the system dynamics, and the same derivation applies to the time-varying $J(u(t),t)$, possibly with stochasticity. In the dynamic case, the perturbation frequency should be chosen to be much higher than the frequency of $J(u^*,t)$, so that the high-pass filtering steps can be effective. Also, the methods can naturally be applied to multi-dimensional control inputs, using different frequencies for each input dimension. The general settings are discussed in detail in~\cite{escbook,zhang2011extremum}. 

\subsection{Comparison with Policy Gradient}

\begin{figure}[t!] 
    \centering
\includegraphics[width=\linewidth]{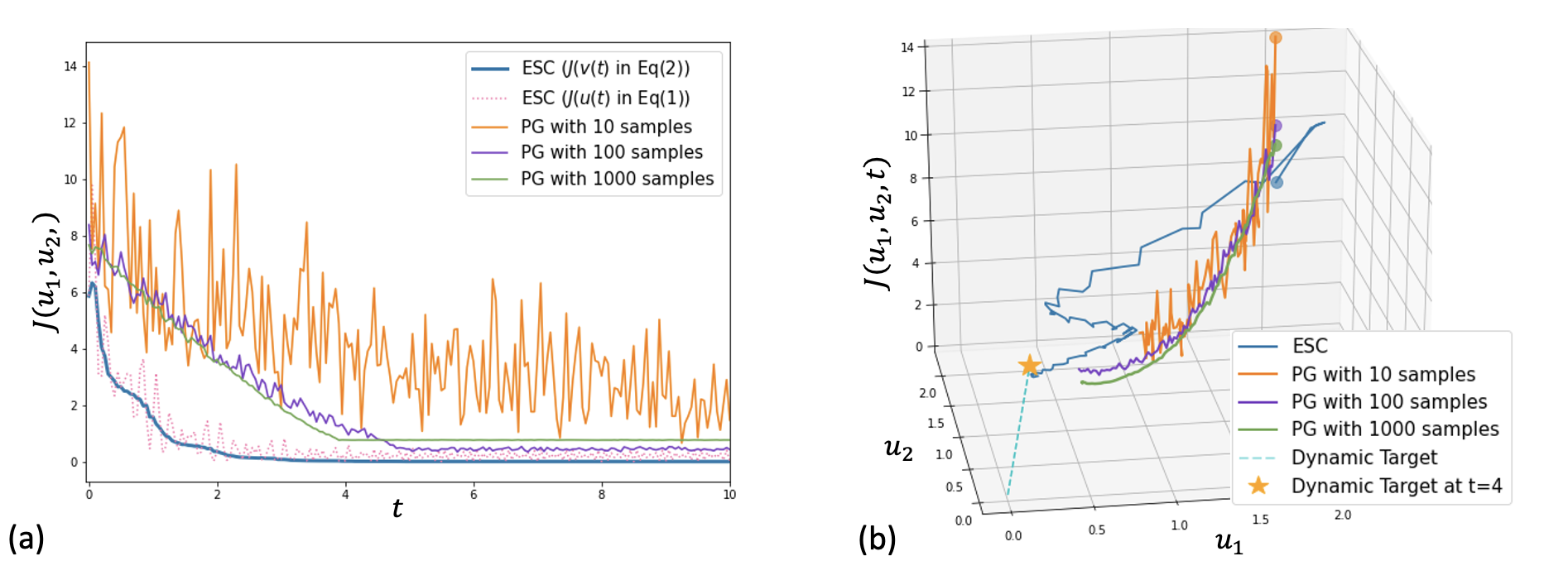}\label{fig:esc}
    \caption{Illustration of the optimum tracking performance between ESC and PG in Example~\ref{ex}. It demonstrates that ESC (blue) exhibits the fastest convergence rate in both static and dynamic optimum examples. (a) The convergence speed in tracking a static objective function. (b) Comparison of the convergence in tracking a time-varying objective function. Trajectories show the convergence towards the optimum over time with varying objective values. The initial point is represented by circle dots, and the goal point at time $t=4$ is denoted by a star.}
    \label{fig:ex} 
\end{figure}

Policy gradient in RL can be considered a special case of the general strategy of search gradient~\cite{JMLR:v15:wierstra14a}, which we can directly compare with ESC. Again consider the setting of optimizing an unknown objective $J(u)$. 
The search gradient approach operates with a parameterized distribution $P_{\theta}(u)$ over the input space with density $p_{\theta}(u)$, and optimizes the following objective in the parameter space $\Theta$:
\begin{equation*}
\max_{\theta\in\Theta}\mathbb{E}_{u\sim P_{\theta}(u)}[J(u)]
\end{equation*}
by iteratively improves the parameters $\theta$ following the gradient of the stochastic objective, which is of the form:
\begin{equation}\label{sg}
\nabla_{\theta}\mathbb{E}_{u\sim P_{\theta}(u)}[J(u)]=\mathbb{E}_{u\sim P_{\theta}(u)}[J(u)\nabla_{\theta}\log(p(u))].
\end{equation}
With appropriate learning rates, following the search gradient ensures convergence to a distribution that centers at a local optimum of the objective $J$. The method benefits from reliable Monte Carlo estimation of the gradient in Eq~(\ref{sg}) with enough samples, which is suitable for offline learning and conservative policy optimization. However, the dynamic updates in ESC can achieve much faster per-sample improvement, as we show in the following example. 
\begin{example}\label{ex}
In Figure~\ref{fig:ex}, we compare ESC and policy gradient on simple objectives in both static and dynamic settings. We first use the static objective $J(u)=(u_1-0.1)^2+(u_2-0.5)^2$ in Figure~\ref{fig:ex}(a), where the initial input is at $(2,2)$. The blue curve at the bottom that shows the fastest convergence to $J(u^*)=0$ is achieved by the estimate $v(t)$ in ESC, and the oscillating dotted curve around it is the response on the actual control input $u(t)$. In contrast, the other curves from policy gradient methods show much slower convergence that is only competitive when 100 samples are used for each update, whereas ESC only queries the objective with one input sample per iteration. In fact, ESC can almost achieve the same progress as gradient descent, which uses the analytic gradient of the function. 
In Figure~\ref{fig:ex}(b), we consider a time-varying objective $J(u,t)=(u_1-0.1t)^2+(u_2-0.5t)^2$. All methods start from $(2,2)$ which is far from the initial optimum of the objective, which is at $(0,0)$. We see that the blue curve representing the estimation with ESC quickly converges to the objective after $t=4$, while policy gradient methods have a hard time quickly tracking the changing objective and the sample size needs to be very large. 
\end{example}
Consequently, it can be beneficial to use ESC to improve the quality of each sample, while maintaining an overall policy gradient framework for reliable improvement. This is the key approach that will be explained in the next section. 

For the RL setting, we use the following standard notations for policy optimization. Markov Decision Processes are tuples $\mathcal M =\langle\mathcal{S}, \mathcal{A}, P, r, \gamma\rangle$ where $\mathcal{S}$ is the state space and $\mathcal{A}$ the action space, and both are continuous in our setting. 
The transition function $P: \mathcal{S} \times \mathcal{S} \times \mathcal{A} \rightarrow [0,1]$ determines the probability $P(s' |s,a) $ of transitioning into state $s'$ from state $s$ after taking action $a$. We consider general forms of reward functions $r: \mathcal{S}\times \mathcal{A}\times \mathcal{S} \rightarrow \mathbb{R}$ defined over transitions, and $\gamma\in [0,1)$ is the discount factor. We write $\pi_\theta$ to denote a stochastic policy $\pi_\theta: \mathcal{S} \times \mathcal{A} \rightarrow [0,1]$ 
parameterized by $\theta$. The goal of policy optimization is to maximize the expected $\gamma$-discounted cumulative return $J_{\rho(s_0)}(\theta) = \mathbb{E}\left[\sum_{t=0}^{\infty} \gamma^t r(s_t, a_t, s_{t+1})\right]$ under some distribution of the initial states $\rho(s_0)$. The $Q$-value of a state-action pair $(s,a)$ under a policy $\pi_{\theta}$ is the expectation of cumulative return of future trajectories after taking $(s,a)$, defined as $Q^{\pi_{\theta}}(s,a)=\mathbb{E}[\sum_{t=0}^{\infty}\gamma^t r(s_t, a_t, s_{t+1})|s_0=s, a_0=a]$. It satisfies the Bellman equation $Q^{\pi_{\theta}}(s,a)=\mathbb{E}[r(s,a,s')+\gamma V^{\pi_{\theta}}(s')]$.

In policy gradient, policy parameters $\theta$ are updated at some learning rate $\alpha$ in the direction of
\[\nabla_{\theta} J(\theta)=\mathbb{E}_{s,a\sim \pi_{\theta}}[A^{\pi_{\theta}}(s,a)\nabla_{\theta}\log \pi_{\theta}(a|s)]\]
where $A^{\pi_{\theta}}(s,a)=Q^{\pi_{\theta}}(s,a)-V^{\pi_{\theta}}(s)$ 
is the advantage of the action $a$ at state $s$, and $\hat{\mathbb{E}}[\cdot]$ denotes the empirical mean estimated through sampled trajectories. In continuous spaces, the behavior policy $\pi_{\theta}(a|s)$ at each state $s$ typically is chosen to be a Gaussian distribution $\mathcal{N}(\mu_s,\Sigma_s)$ over the action space. 

\section{Extremum-Seeking Action Selection}
\label{sec:esa}

We now describe the Extreme-Seeking Action Selection (ESA) method, which uses ESC strategies for improving each action sample to attain higher advantages locally, with exploration driven by both the sampling distribution from the behavior policy and the perturbations within ESC. The overall algorithm is shown in Algorithm~\ref{alg:pes}. 

ESA can be used as an add-on component for improving the quality of each action sample in typical policy optimization algorithms. We write $a_t\sim \pi_{\theta}(\cdot|s_t)$ as the action sample drawn at state $s_t$ at time step $t$, according to the distribution determined by the current policy $\pi_{\theta}$. Following ESC design in Equation (\ref{first}) and (\ref{eq:v}), we need to maintain two vectors: $u(t)$ as the oscillatory control input, and $v(t)$ for updating an estimation that approaches the optimum. 
In the RL setting, persistent perturbation on the action hinders convergence and policy improvement. So instead of using oscillatory inputs as actions, we use $u(t)$ to probe estimations of Q-values, from which we find a reliable improvement $a_t+v(t)$ of the original action sample, which is then applied to the environment. $u(0)$, $v(0)$, and $t$ are all set to zero in the beginning of each episode. 

Concretely, as shown in Algorithm~\ref{alg:pes}, for each sampled action $a_t$, we first apply sinusoidal perturbation as $a_t+u(t)$, where $u(t)=v(t)+K\sin(\omega t)$. We use the $Q$-values determined by the current policy $\pi_{\theta}$ as the objective, and query for the value of $Q(s,a_t+u(t))$. Next, we use a modification of Equation (\ref{eq:v}) to update $v(t)$ as follows:
\begin{equation}\label{eq:vupdate}
v(t+1) \gets v(t)+\alpha \sin(\omega t)\mathbb{H}[Q(s,a_t+u(t))]    
\end{equation}
where $\mathbb{H}$ is a high-pass filter, and $\alpha>0$ is a learning rate. We do not need the minus sign in Equation (\ref{eq:v}), which was for minimizing the objective and here we maximize. Importantly, we have also dropped the low-pass filter $\mathbb{L}[\cdot]$ from the standard design of ESC, because we want to still allow some high-frequency perturbations for enabling exploration, which is different from the ESC goal of tracking the extremum as precisely as possible. In this way, exploration of actions is achieved first by the original sample $a_t\sim \pi_{\theta}(\cdot|s_t)$, and then with sinusoidal components in $v(t)$ from the definition above. Finally, $a_t+v(t)$ will be the actual action applied to the environment, and we move onto the next state $s_{t+1}$ by querying the environment with $(s_t, a_t+v(t))$. 

We take advantage of the ability of ESC methods for directly tracking time-varying objectives, which is the Q-values that change over time steps within each episode. 
It bypasses a major challenge for exploration in continuous spaces, where we can not easily keep track of properties of individual states (such as visitation counts that are often used for exploration in discrete spaces). Instead, using the control-theoretic and frequency-domain analysis, we shift the focus to time-varying perturbation throughout the entire trajectories to achieve improvement in the overall performance. 

\begin{figure}[ht] 
    \centering
    \includegraphics[width=\linewidth]{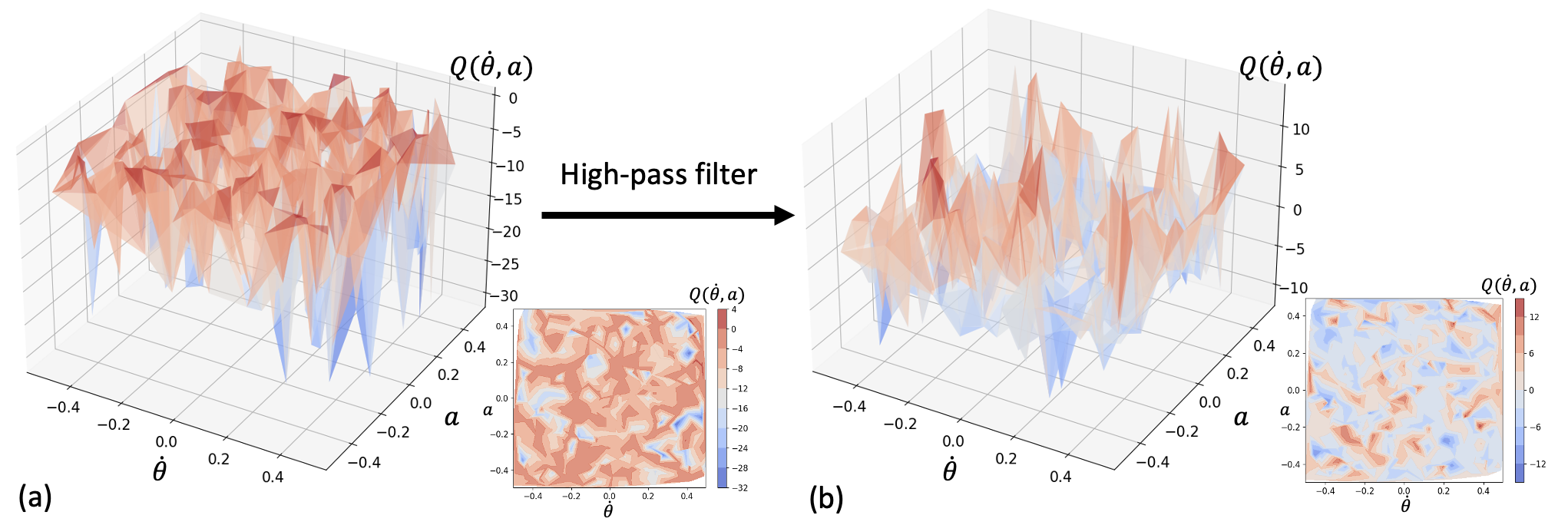}
    \caption{An illustration of the effect of using high-pass filters on the Q-value landscapes. (a) A Q-value landscape at a state in the inverted pendulum environment, plotted for a fixed policy $\pi_{\theta}$ at an intermediate stage of training. (b) Filtered Q-value landscape from (a).} 
    \label{fig:Q}
\end{figure} 

\begin{algorithm}[h!t]
  \caption{Policy Optimization with ESA}\label{alg:pes}
  \begin{algorithmic}[1]
       \State Randomly initialize the policy network parameters $\theta$ and Q-network parameters $\phi$, and empty replay buffer $\mathcal{D}$. 
       \State Choose hyperparameters: perturbation amplitudes vector $K$, frequency vector $\omega$, and learning rate vector $\alpha$,
       \For{episodes $= 1,\dots,N$} 
           \State $u(0) \gets 0$ and $v(0) \gets 0$
           \For{$t = 0,\dots,T$}
               \State Sample $a_t \sim \pi_{\theta}(\cdot|s_t)$
               % \IF{episodes <= $E$}
               \State $u(t) \gets v(t) + K\sin{\omega t}$ 
               \Comment{Following Eq~(\ref{first})}
               \State $v(t+1) \gets v(t)+\alpha K\sin(\omega t)\mathbb{H}[Q(s,a_t+u(t))]$\\       \Comment{Filter based on $Q$-values, as Eq. (\ref{eq:vupdate})}
               \State $s_{t+1}\gets \mathrm{Env}(s_t, a_t+v(t))$
               %\Comment{Sample next state from the environment}
               \State $\mathcal{D} \leftarrow \mathcal{D} \cup {(s_t, a_t+v(t), r_t, s_{t+1})}$ 
           \EndFor
           \For {each policy optimization step}
               % \State Sample mini-batches of transitions from $\mathcal{D}$
               \State Update $\theta$ and $\phi$ with $\mathcal{D}$ using standard policy optimization algorithms
          \EndFor     
       \EndFor
\end{algorithmic}
\end{algorithm}

\noindent{\bf Hyperparameters.} The new hyperparameters introduced by the ESA component include the perturbation magnitude vector $K$, sinusoidal perturbation frequency vector $\omega$, and the learning rate vector $\alpha$. In particular, as long as the perturbation frequency is reasonably higher than the frequency of the Q-value function, the high-pass filter will be able to isolate the local second-order information of the objective. Effects of the hyperparameter choices will be further discussed through ablation study in Section~\ref{sec:exp}.

\noindent\textbf{Benefits of High-Pass Filtering.} High-pass filtering is an important step in ESC that ensures the convergence of the design of the algorithm. Intuitively, in the context of action selection, high-pass filters remove ``flat'' regions in the Q-value landscape, making it easier to locate actions that lead to local peak Q-values. In Figure~\ref{fig:Q}, we observe that high-pass filters enhance the visibility of peaks and enable faster local improvement towards the optimum.

\begin{figure*}[h!]    
    \centering
\includegraphics[width=\textwidth]{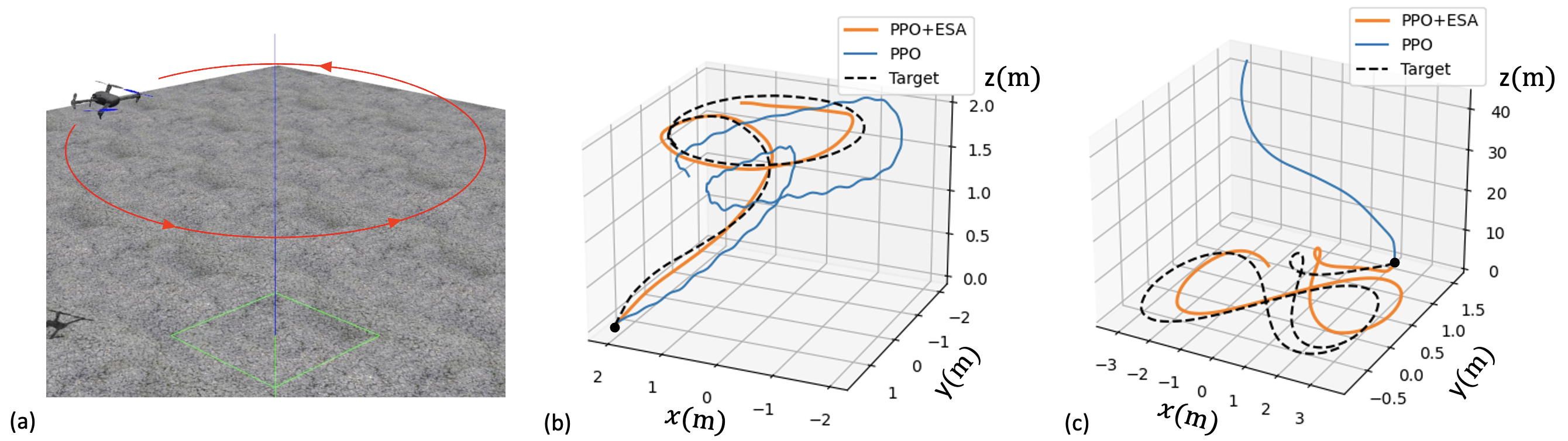}
    \caption{Illustration of how ESA improves the performance of PPO for quadrotor environment. We evaluated both policies trained after the same number of iterations. We observe that ESA improves the quality of sampled actions and accelerates learning. (a) Quadrotor control environment. (b) Performance comparison in a circle target path task. (c) Performance comparison in tracking an eight-shaped target path, where the PPO-trained policy diverges.}
    \label{fig:quad_path}
\end{figure*}

\begin{figure*}[ht]    
    \centering
    \includegraphics[width=0.49\linewidth]{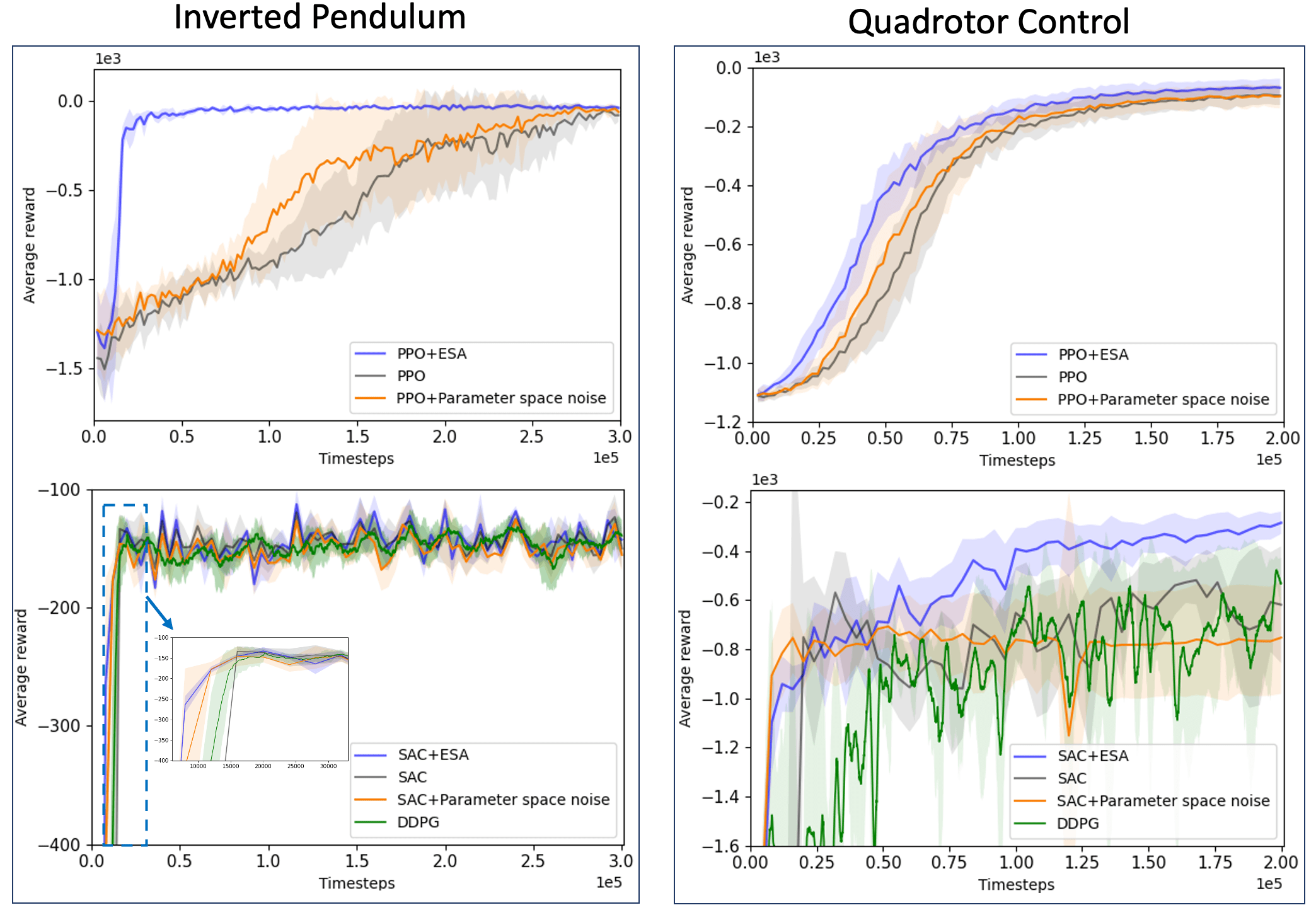}
    \includegraphics[width=0.49\linewidth]{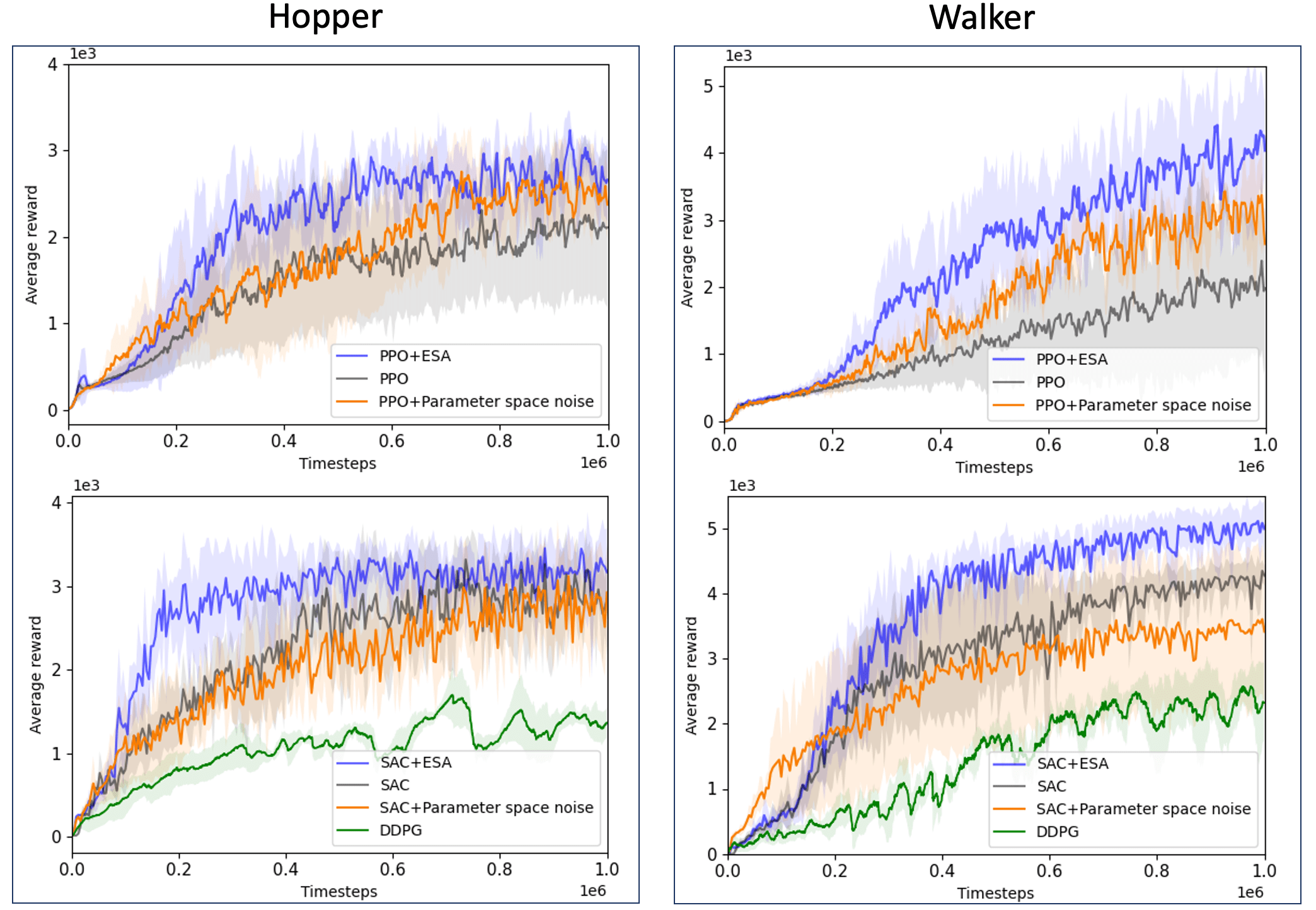}
    \caption{Performance comparison for all methods. PPO+ESA (blue, first row) and SAC+ESA (blue, second row) demonstrate higher learning efficiency and performance
    compared to other methods across all tasks. In comparison, adding random parameter noise (orange) leads to better exploration in the early stages of some tasks, but fails to sustain effective exploration throughout the entire training process.}
    \label{fig:lc_mujoco}
\end{figure*}

\noindent\textbf{Comparison with Using Analytic Gradient of $Q$-Networks.} A natural approach of improving action samples to higher quality is to follow the gradient of the $Q$-value networks, i.e., using a first-order approach rather than the zeroth-order one proposed here. Namely, for each state $s_t$ and sampled action $a_t$, we can query the $Q$-network for its gradient at $(s_t,a_t)$ over actions, which in principle should indicate the direction of moving the sampled action towards higher $Q$-values. However, the $Q$-value models provided by deep neural networks have highly non-smooth landscapes over action inputs, as illustrated in Figure~\ref{fig:Q}. Thus the analytic gradients are frequently misleading. Instead, ESC provides robust estimation through filtering and frequency analysis. 

\section{Experiments}\label{sec:exp}

We show experimental results to evaluate how ESA can improve the performance of policy optimization. We add ESA to the leading policy optimization methods including Proximal Policy Optimization (PPO)~\cite{ppo} and Soft Actor-Critic (SAC)~\cite{sac} and benchmark the performance difference in various challenging control learning environments. 

\noindent\textbf{Environments.} We consider continuous control environments in OpenAI Gym~\cite{OpenAI} and MuJoCo~\cite{MuJoCo}, including the inverted pendulum, hopper, and walker, as well as a Gazebo-based quadrotor control simulator enabled by the commercially-used autopilot framework PX4~\cite{meier2015px4}. The quadrotor control environment involves 12 state dimensions (inertia frame positions, velocities, rotation angles, and angular velocities) and 4 control inputs (thrust, roll, pitch, and yaw). Details of the equations of motion of the quadrotor can be found in~\cite{UAVsurvey}. The goal of the agent is to track an oriented point along a path, and the rewards are calculated based on the discrepancies between their positions and orientations.

\noindent\textbf{Baselines.} We compare the performance of PPO+ESA and SAC+ESA with the standard PPO and SAC, as well as with the versions using additional parameter space noise, a widely-used approach for enhancing exploration~\cite{DBLP:conf/iclr/PlappertHDSC0AA18}. We also show compare with DDPG incorporating time-correlated Ornstein–Uhlenbeck noise~\cite{ddpg}. All algorithms are tested on 5 different random seeds in all environments.

\noindent\textbf{Overall Performance.} Figure \ref{fig:lc_mujoco} shows comparisons of learning curves for all methods in benchmark environments. We observe that ESA accelerates learning and enhances the performance of both PPO and SAC, and outperforms other baselines. The computational cost of adding ESA is at most 50 percent longer runtime for each episode (2048 steps). In particular, Figure~\ref{fig:quad_path} demonstrates the specific improvement in performance in the quadrotor control environment. We visualize the behaviors of the trained control policies after the same number of training steps, and observe that PPO+ESA shows clear improvement in the control performance compared to the original PPO-trained policy. 

\begin{figure*}[h!]    
    \centering
\includegraphics[width=\linewidth]{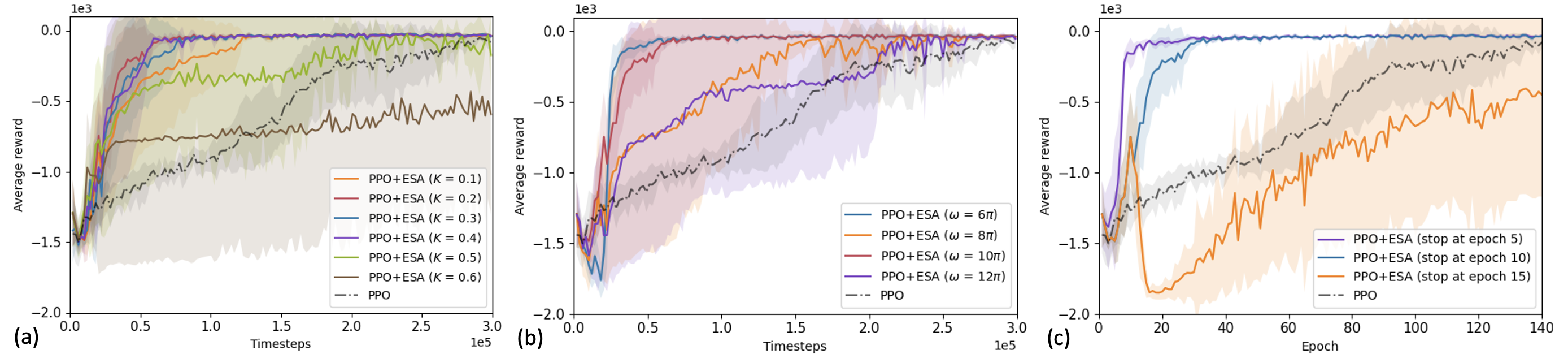}
    \caption{Ablation studies conducted on the inverted pendulum environment. (a) Training curves in relation to the magnitude of the perturbation signal. (b) Training curves in relation to the frequency of the perturbation signal. (c) Training curves comparing different numbers of ESA intervening episodes.}
    \label{fig:ablation}
\end{figure*}

\noindent\textbf{Ablation Study: Perturbation Magnitude.} 
The parameter amplitude $K$ of the perturbation signal presents a trade-off between increasing convergence speed and reducing oscillation. Figure \ref{fig:ablation}(a) shows how the learning performance changes as at various values of $K$ for the perturbation signal in the inverted pendulum environment, when the frequency of the perturbation is fixed. We see that there the magnitude of $K=0.2$ (red) achieves the best outcome. Reducing $K$ to $0.1$ leads to a slower convergence speed. Increasing $K$ to be above $0.2$ accelerates the initial progress of the learning curves but results in much higher variance in performance across different random seeds.

\vspace{1mm}
\noindent\textbf{Ablation Study: Perturbation Frequency.} 
The effective perturbation frequencies are affected by the environment dynamics and the responsiveness of the Q-value functions. In general, higher $\omega$ allows us to obtain a more accurate gradient estimate by applying a high-pass filter to the value. However, very high frequency may lead to non-smooth action choices that negatively impact policy learning. On the other hand, as shown in Figure \ref{fig:ablation}(b), we observe that when the frequecy gets higher than $10\pi$ the effectiveness of ESA is reduced. 

\vspace{1mm}
\noindent\textbf{Ablation Study: Decay of ESA Learning Rate.} 
In the policy optimization process, when the policy has reached near-peak performance, it also becomes very sensitive to perturbations. Thus the ESA-driven exploration should decay over time to avoid destabilizing policy learning. The results in Figure \ref{fig:ablation}(c) illustrate the impact of different decay rates on performance.  

\section{Conclusion}

We proposed the extremum-seeking action selection (ESA) method for improving both exploration and exploitation in sampling actions for policy optimization in continuous spaces. We follow the strategies in extremum-seeking control (ESC) by applying sinusoidal perturbations on the sampled actions in each step to obtain actions of higher action values and also improve exploration. We have shown that ESC methods can be particularly sample efficient for locally optimizing unknown objectives, compared to policy gradient methods. At the same time, the scale of ESA perturbations on the sampled actions needs to be carefully chosen to balance the trade-off between fast local improvement with ESC and reliable policy improvement over all states. The ability of tracking dynamic objectives makes ESC methods particularly suitable for handling problems in the continuous domain by shifting the focus from states to improving entire trajectories over time. We observed clear benefits of adding ESA methods in PPO and SAC in improving the learning performance in various continuous control problems. 

\section*{Appendix: More Details in the Derivation of ESC}
Because ESC techniques have rarely been introduced in the context of reinforcement learning, we provide more derivation details on how ESC ensures that its estimate $v(t)$ converges locally to an optimum, following Equation~(\ref{first}-\ref{eq:v})
We use the one-dimensional input for simplicity, and more general derivations can be found in standard references such as~\cite{escbook}. 

Consider minimizing $J(u)$ near a strict local minimizer $u^*$, which means that the first-order derivative $J'(u^*)=0$, and the second-order derivative $J''(u^*)>0$ in a local neighborhood of $u^*$. Since the analysis is local, we can approximate $J(u(t))$ with the second-order Taylor expansion:
\small
\begin{eqnarray*}
J(u(t))\hspace{-3mm} &\cong& \hspace{-3mm} J(u^*)+\frac{1}{2}J''(u^*)(u(t)-u^*)^2\\
&=&\hspace{-3mm} J(u^*) + \frac{1}{2}J''(u^*)(K\sin (\omega t)+v(t)-u^*)^2
\end{eqnarray*}
where the first-order term $J'(u^*)=0$, and the second equality is by plugging in $u(t)$ from Equation (\ref{first}). Now, a key step is to focus on the error dynamics, which is how the difference between $v(t)$ and $u^*$ changes, by defining $\xi(t)=u^*-v(t)\label{xi}$. The previous expansion can then be rewritten as:
\begin{multline*}
J(u(t)) 
= J(u^*)+\frac{1}{2}J''(u^*)\xi(t)^2-KJ''(u^*)\sin (\omega t) \xi(t) \\
+ \frac{1}{4}K^2J''(u^*)(1-\cos 2\omega t)
\end{multline*}
where we use $\sin^2(\omega t)=\frac{1}{2}(1-\cos 2\omega t)$. Now, by applying a high-frequency filter $\mathbb{H}$ on $J(u(t))$, we can remove the terms at lower frequencies such as $J(u^*)$ and $\frac{1}{4}K^2J''(u^*)$. We also remove the second-order term $\xi(t)^2$ dominated by $\xi(t)$ in the local analysis:
\begin{equation*}
\mathbb{H}[J(u(t))] \cong -KJ''(u^*)\sin (\omega t)\xi(t) - \frac{1}{2}K^2J''(u^*)\cos (2\omega t) 
\end{equation*}
Here we see why the filtering mechanism is useful: it allows us to inspect the signal only at certain frequencies that carry the information we need, in this case $J''(u^*)$. To fully do that, we will further demodulate the signal with $\sin(\omega t)$ and then apply low-pass filtering and arrive at the approximation:
\begin{equation*}
\mathbb{L}[\sin(\omega t)\mathbb{H}[J(u(t))]]\cong -\frac{1}{2}KJ''(u^*)\xi(t)
\end{equation*}
where the sinusoidal terms are all filtered out by the low-pass filter $\mathbb{L}[\cdot]$ because of their high frequency. Now, plugging it in the definition of $\xi(t)$ and $\dot v(t)$ from Equation, we arrive at Equation~\ref{eq:third} in Section~\ref{sec:esc}. Consequently, the error dynamics $\xi(t)$ follows exponentially stabilizes to zero. Through such frequency-domain analysis, we see that $J''(u^*)$ does not need to be estimated, and convergence is implicitly guaranteed by following the control law for $u(t)$ and $v(t)$. 

\newpage
\bibliographystyle{IEEEtran}
\bibliography{refs}

% Generated by IEEEtran.bst, version: 1.14 (2015/08/26)
\begin{thebibliography}{10}
\providecommand{\url}[1]{#1}
\csname url@samestyle\endcsname
\providecommand{\newblock}{\relax}
\providecommand{\bibinfo}[2]{#2}
\providecommand{\BIBentrySTDinterwordspacing}{\spaceskip=0pt\relax}
\providecommand{\BIBentryALTinterwordstretchfactor}{4}
\providecommand{\BIBentryALTinterwordspacing}{\spaceskip=\fontdimen2\font plus
\BIBentryALTinterwordstretchfactor\fontdimen3\font minus \fontdimen4\font\relax}
\providecommand{\BIBforeignlanguage}[2]{{%
\expandafter\ifx\csname l@#1\endcsname\relax
\typeout{** WARNING: IEEEtran.bst: No hyphenation pattern has been}%
\typeout{** loaded for the language `#1'. Using the pattern for}%
\typeout{** the default language instead.}%
\else
\language=\csname l@#1\endcsname
\fi
#2}}
\providecommand{\BIBdecl}{\relax}
\BIBdecl

\bibitem{PengCZLTL20}
X.~B. Peng, E.~Coumans, T.~Zhang, T.~E. Lee, J.~Tan, and S.~Levine, ``Learning agile robotic locomotion skills by imitating animals,'' in \emph{Robotics: Science and Systems XVI, Virtual Event / Corvalis, Oregon, USA, July 12-16, 2020}, M.~Toussaint, A.~Bicchi, and T.~Hermans, Eds., 2020.

\bibitem{kalashnikov18a}
D.~Kalashnikov, A.~Irpan, P.~Pastor, J.~Ibarz, A.~Herzog, E.~Jang, D.~Quillen, E.~Holly, M.~Kalakrishnan, V.~Vanhoucke, and S.~Levine, ``Scalable deep reinforcement learning for vision-based robotic manipulation,'' in \emph{Proceedings of The 2nd Conference on Robot Learning}, 2018.

\bibitem{wang2019autonomous}
C.~Wang, J.~Wang, Y.~Shen, and X.~Zhang, ``Autonomous navigation of uavs in large-scale complex environments: A deep reinforcement learning approach,'' \emph{IEEE Transactions on Vehicular Technology}, vol.~68, no.~3, pp. 2124--2136, 2019.

\bibitem{li2021reinforcement}
Z.~Li, X.~Cheng, X.~B. Peng, P.~Abbeel, S.~Levine, G.~Berseth, and K.~Sreenath, ``Reinforcement learning for robust parameterized locomotion control of bipedal robots,'' in \emph{2021 IEEE International Conference on Robotics and Automation (ICRA)}.\hskip 1em plus 0.5em minus 0.4em\relax IEEE, 2021, pp. 2811--2817.

\bibitem{KaufmannLR0K020}
E.~Kaufmann, A.~Loquercio, R.~Ranftl, M.~M{\"{u}}ller, V.~Koltun, and D.~Scaramuzza, ``Deep drone acrobatics,'' in \emph{Robotics: Science and Systems XVI, Virtual Event / Corvalis, Oregon, USA, July 12-16, 2020}, M.~Toussaint, A.~Bicchi, and T.~Hermans, Eds., 2020.

\bibitem{chang2021stabilizing}
Y.-C. Chang and S.~Gao, ``Stabilizing neural control using self-learned almost lyapunov critics,'' in \emph{2021 IEEE International Conference on Robotics and Automation (ICRA)}.\hskip 1em plus 0.5em minus 0.4em\relax IEEE, 2021.

\bibitem{so2023solving}
O.~So and C.~Fan, ``Solving stabilize-avoid optimal control via epigraph form and deep reinforcement learning,'' in \emph{Proceedings of Robotics: Science and Systems}, 2023.

\bibitem{ganai2023iterative}
M.~Ganai, Z.~Gong, C.~Yu, S.~L. Herbert, and S.~Gao, ``Iterative reachability estimation for safe reinforcement learning,'' in \emph{Thirty-seventh Conference on Neural Information Processing Systems}, 2023.

\bibitem{trpo}
J.~Schulman, S.~Levine, P.~Abbeel, M.~I. Jordan, and P.~Moritz, ``Trust region policy optimization,'' in \emph{ICML 2015}, 2015, pp. 1889--1897.

\bibitem{ddpg}
T.~P. Lillicrap, J.~J. Hunt, A.~Pritzel, N.~Heess, T.~Erez, Y.~Tassa, D.~Silver, and D.~Wierstra, ``Continuous control with deep reinforcement learning,'' \emph{arXiv preprint arXiv:1509.02971}, 2015.

\bibitem{ppo}
J.~Schulman, F.~Wolski, P.~Dhariwal, A.~Radford, and O.~Klimov, ``Proximal policy optimization algorithms,'' \emph{arXiv preprint arXiv:1707.06347}, 2017.

\bibitem{td3}
S.~Fujimoto, H.~Hoof, and D.~Meger, ``Addressing function approximation error in actor-critic methods,'' in \emph{International conference on machine learning}.\hskip 1em plus 0.5em minus 0.4em\relax PMLR, 2018, pp. 1587--1596.

\bibitem{sac}
T.~Haarnoja, A.~Zhou, P.~Abbeel, and S.~Levine, ``Soft actor-critic: Off-policy maximum entropy deep reinforcement learning with a stochastic actor,'' in \emph{Proceedings of the 35th International Conference on Machine Learning}, ser. Proceedings of Machine Learning Research, vol.~80.\hskip 1em plus 0.5em minus 0.4em\relax Stockholmsmässan, Stockholm Sweden: PMLR, 10--15 Jul 2018, pp. 1861--1870.

\bibitem{engstrom2020implementation}
L.~Engstrom, A.~Ilyas, S.~Santurkar, D.~Tsipras, F.~Janoos, L.~Rudolph, and A.~Madry, ``Implementation matters in deep policy gradients: A case study on ppo and trpo,'' \emph{arXiv preprint arXiv:2005.12729}, 2020.

\bibitem{hsu2020revisiting}
C.~C.-Y. Hsu, C.~Mendler-D{\"u}nner, and M.~Hardt, ``Revisiting design choices in proximal policy optimization,'' \emph{arXiv preprint arXiv:2009.10897}, 2020.

\bibitem{ibarz2021train}
J.~Ibarz, J.~Tan, C.~Finn, M.~Kalakrishnan, P.~Pastor, and S.~Levine, ``How to train your robot with deep reinforcement learning: lessons we have learned,'' \emph{The International Journal of Robotics Research}, vol.~40, no. 4-5, pp. 698--721, 2021.

\bibitem{hu2020learning}
Y.~Hu, W.~Wang, H.~Jia, Y.~Wang, Y.~Chen, J.~Hao, F.~Wu, and C.~Fan, ``Learning to utilize shaping rewards: A new approach of reward shaping,'' \emph{Advances in Neural Information Processing Systems}, 2020.

\bibitem{gupta2022unpacking}
A.~Gupta, A.~Pacchiano, Y.~Zhai, S.~Kakade, and S.~Levine, ``Unpacking reward shaping: Understanding the benefits of reward engineering on sample complexity,'' \emph{Advances in Neural Information Processing Systems}, vol.~35, pp. 15\,281--15\,295, 2022.

\bibitem{escbook}
K.~B. Ariyur and M.~Krstic, \emph{Real Time Optimization by Extremum Seeking Control}.\hskip 1em plus 0.5em minus 0.4em\relax USA: John Wiley \& Sons, Inc., 2003.

\bibitem{zhang2011extremum}
C.~Zhang and R.~Ord{\'o}{\~n}ez, \emph{Extremum-seeking control and applications: a numerical optimization-based approach}.\hskip 1em plus 0.5em minus 0.4em\relax Springer Science \& Business Media, 2011.

\bibitem{nevsic2009extremum}
D.~Ne{\v{s}}i{\'c}, ``Extremum seeking control: Convergence analysis,'' \emph{European Journal of Control}, vol.~15, no. 3-4, pp. 331--347, 2009.

\bibitem{DBLP:conf/nips/0002GBB18}
W.~Sun, G.~J. Gordon, B.~Boots, and J.~A. Bagnell, ``Dual policy iteration,'' in \emph{Advances in Neural Information Processing Systems 31: Annual Conference on Neural Information Processing Systems 2018, NeurIPS 2018, December 3-8, 2018, Montr{\'{e}}al, Canada}, S.~Bengio, H.~M. Wallach, H.~Larochelle, K.~Grauman, N.~Cesa{-}Bianchi, and R.~Garnett, Eds., 2018, pp. 7059--7069.

\bibitem{MuJoCo}
E.~Todorov, T.~Erez, and Y.~Tassa, ``Mujoco: A physics engine for model-based control,'' in \emph{2012 IEEE/RSJ International Conference on Intelligent Robots and Systems}, 2012, pp. 5026--5033.

\bibitem{khan2012reinforcement}
S.~G. Khan, G.~Herrmann, F.~L. Lewis, T.~Pipe, and C.~Melhuish, ``Reinforcement learning and optimal adaptive control: An overview and implementation examples,'' \emph{Annual reviews in control}, vol.~36, no.~1, pp. 42--59, 2012.

\bibitem{richardsadaptive}
S.~M. Richards, N.~Azizan, J.-J. Slotine, and M.~Pavone, ``Adaptive-control-oriented meta-learning for nonlinear systems.''

\bibitem{westenbroek2020adaptive}
T.~Westenbroek, E.~Mazumdar, D.~Fridovich-Keil, V.~Prabhu, C.~J. Tomlin, and S.~S. Sastry, ``Adaptive control for linearizable systems using on-policy reinforcement learning,'' in \emph{2020 59th IEEE Conference on Decision and Control (CDC)}.\hskip 1em plus 0.5em minus 0.4em\relax IEEE, 2020, pp. 118--125.

\bibitem{annaswamy2023integration}
A.~M. Annaswamy, A.~Guha, Y.~Cui, S.~Tang, P.~A. Fisher, and J.~E. Gaudio, ``Integration of adaptive control and reinforcement learning for real-time control and learning,'' \emph{IEEE Transactions on Automatic Control}, 2023.

\bibitem{wang2023reinforcement}
Y.~Wang, C.~Zheng, M.~Sun, Z.~Chen, and Q.~Sun, ``Reinforcement-learning-aided adaptive control for autonomous driving with combined lateral and longitudinal dynamics,'' in \emph{2023 IEEE 12th Data Driven Control and Learning Systems Conference (DDCLS)}.\hskip 1em plus 0.5em minus 0.4em\relax IEEE, 2023, pp. 840--845.

\bibitem{Survey_exploration}
S.~Amin, M.~Gomrokchi, H.~Satija, H.~van Hoof, and D.~Precup, ``A survey of exploration methods in reinforcement learning,'' \emph{CoRR}, vol. abs/2109.00157, 2021.

\bibitem{wawrzynski2015control}
P.~Wawrzynski, ``Control policy with autocorrelated noise in reinforcement learning for robotics,'' \emph{International Journal of Machine Learning and Computing}, vol.~5, no.~2, p.~91, 2015.

\bibitem{eberhard2022pink}
O.~Eberhard, J.~Hollenstein, C.~Pinneri, and G.~Martius, ``Pink noise is all you need: Colored noise exploration in deep reinforcement learning,'' in \emph{The Eleventh International Conference on Learning Representations}, 2022.

\bibitem{ruckstiess2010exploring}
T.~R{\"u}ckstiess, F.~Sehnke, T.~Schaul, D.~Wierstra, Y.~Sun, and J.~Schmidhuber, ``Exploring parameter space in reinforcement learning,'' \emph{Paladyn}, vol.~1, no.~1, pp. 14--24, 2010.

\bibitem{conti2018improving}
E.~Conti, V.~Madhavan, F.~P. Such, J.~Lehman, K.~O. Stanley, and J.~Clune, ``Improving exploration in evolution strategies for deep reinforcement learning via a population of novelty-seeking agents,'' in \emph{Advances in Neural Information Processing Systems}, 2018.

\bibitem{fortunato2018noisy}
\BIBentryALTinterwordspacing
M.~Fortunato, M.~G. Azar, B.~Piot, J.~Menick, M.~Hessel, I.~Osband, A.~Graves, V.~Mnih, R.~Munos, D.~Hassabis, O.~Pietquin, C.~Blundell, and S.~Legg, ``Noisy networks for exploration,'' in \emph{International Conference on Learning Representations}, 2018. [Online]. Available: \url{https://openreview.net/forum?id=rywHCPkAW}
\BIBentrySTDinterwordspacing

\bibitem{DBLP:conf/iclr/PlappertHDSC0AA18}
\BIBentryALTinterwordspacing
M.~Plappert, R.~Houthooft, P.~Dhariwal, S.~Sidor, R.~Y. Chen, X.~Chen, T.~Asfour, P.~Abbeel, and M.~Andrychowicz, ``Parameter space noise for exploration,'' in \emph{6th International Conference on Learning Representations, {ICLR} 2018, Vancouver, BC, Canada, April 30 - May 3, 2018, Conference Track Proceedings}.\hskip 1em plus 0.5em minus 0.4em\relax OpenReview.net, 2018. [Online]. Available: \url{https://openreview.net/forum?id=ByBAl2eAZ}
\BIBentrySTDinterwordspacing

\bibitem{mahajan2019maven}
A.~Mahajan, T.~Rashid, M.~Samvelyan, and S.~Whiteson, ``Maven: Multi-agent variational exploration,'' in \emph{Advances in Neural Information Processing Systems}, 2019, pp. 7611--7622.

\bibitem{JMLR:v15:wierstra14a}
\BIBentryALTinterwordspacing
D.~Wierstra, T.~Schaul, T.~Glasmachers, Y.~Sun, J.~Peters, and J.~Schmidhuber, ``Natural evolution strategies,'' \emph{Journal of Machine Learning Research}, vol.~15, no.~27, pp. 949--980, 2014. [Online]. Available: \url{http://jmlr.org/papers/v15/wierstra14a.html}
\BIBentrySTDinterwordspacing

\bibitem{OpenAI}
\BIBentryALTinterwordspacing
G.~Brockman, V.~Cheung, L.~Pettersson, J.~Schneider, J.~Schulman, J.~Tang, and W.~Zaremba, ``Openai gym,'' 2016. [Online]. Available: \url{https://arxiv.org/abs/1606.01540}
\BIBentrySTDinterwordspacing

\bibitem{meier2015px4}
L.~Meier, D.~Honegger, and M.~Pollefeys, ``Px4: A node-based multithreaded open source robotics framework for deeply embedded platforms,'' in \emph{2015 IEEE international conference on robotics and automation (ICRA)}.\hskip 1em plus 0.5em minus 0.4em\relax IEEE, 2015, pp. 6235--6240.

\bibitem{UAVsurvey}
B.~Rubí, R.~Pérez, and B.~Morcego, ``A survey of path following control strategies for uavs focused on quadrotors,'' \emph{Journal of Intelligent \& Robotic Systems}, 05 2020.

\end{thebibliography}

\end{document}